\documentclass{article}

\usepackage[utf8]{inputenc} 
\usepackage[T1]{fontenc}    
\usepackage{hyperref}       
\usepackage{url}            
\usepackage{booktabs}       
\usepackage{amsfonts}       
\usepackage{nicefrac}       
\usepackage{microtype}      

\usepackage{algorithm, algpseudocode}  
\usepackage{graphicx}
\usepackage{subcaption}
\usepackage{amsmath,amssymb,amsthm}
\usepackage{appendix}
\newtheorem{theorem}{Theorem}[subsection]

\newtheorem{corollary}{Corollary}[theorem]

\usepackage{color}
\usepackage{dsfont}
\usepackage{xr}

\begin{document}
\title{GAN-EM: GAN based EM Learning Framework}

%

\author{
  Wentian Zhao$^1$, Shaojie Wang$^1$, Zhihuai Xie$^2$, Jing Shi$^1$,\\
  Chenliang Xu$^1$\\
  $^1$University of Rochester, $^2$Tsinghua University
}
\date{}
\maketitle
\begin{abstract}
Expectation maximization (EM) algorithm is to find maximum likelihood solution for models having latent variables.
A typical example is Gaussian Mixture Model (GMM) which requires Gaussian assumption, however, natural images are highly non-Gaussian so that GMM cannot be applied to perform image clustering task on pixel space.
To overcome such limitation, we propose a GAN based EM learning framework that can maximize the likelihood of images and estimate the latent variables with only the constraint of L-Lipschitz continuity.
We call this model GAN-EM, which is a framework for image clustering, semi-supervised classification and dimensionality reduction.
In M-step, we design a novel loss function for discriminator of GAN to perform maximum likelihood estimation (MLE) on data with soft class label assignments. Specifically, a conditional generator captures data distribution for $K$ classes, and a discriminator tells whether a sample is real or fake for each class.
Since our model is unsupervised, the class label of real data is regarded as latent variable, which is estimated by an additional network (E-net) in E-step.
The proposed GAN-EM achieves state-of-the-art clustering and semi-supervised classification results on MNIST, SVHN and CelebA, as well as comparable quality of generated images to other recently developed generative models.
\end{abstract}

\section{Introduction}
\label{sec:intro}

Expectation maximization (EM)~\cite{EM} is a traditional learning framework, which has various applications in unsupervised learning. A typical example is Gaussian mixture model (GMM), where data distribution is estimated by maximum likelihood estimate (MLE) under the Gaussian assumption in M-step, and soft class labels are assigned using Bayes rule in E-step. Although GMM has the nice property that likelihood increases monotonically, many previous studies~\cite{non_Gaussian1}~\cite{non_Gaussian2}~\cite{non_Gaussian3} have shown that natural image intensities exhibit highly non-Gaussian behaviors so that GMM cannot be applied to image clustering on pixel space directly. The motivation of this work is to find an alternative way to achieve EM mechanism without Gaussian assumption. 

Generative adversarial network (GAN)~\cite{GAN} has been proved to be powerful on learning data distribution. We propose to apply it in M-step to maximize the likelihood of data with soft class label assignments passed from E-step. It is easy for GAN to perform MLE as in~\cite{gantutorial,f-gan}, but to incorporate the soft class assignments into the GAN model in the meantime is rather difficult. To address this problem, we design a weighted binary cross entropy loss function for discriminator where the weights are the soft label assignments. In Sec.~\ref{sec:theory}, we prove that such design enables GAN to optimize the Q function of EM algorithm. Since neural networks are not reversible in most cases, we could not use Bayes rule to compute the expectation analytically in E-step like that for GMM. To solve this, we use generated samples to train another network, named E-net, then predict the soft class labels for real samples in E-step.


To evaluate our model, we perform the clustering task based on MNIST and achieve lowest error rate with all 3 different numbers of clusters: 10, 20 and 30, which are common settings in previous works. We also test the semi-supervised classification performance on MNIST and SVHN with partially labeled data, both results being rather competitive compared to recently proposed generative models. Especially, on SVHN dataset, GAN-EM outperforms all other models. Apart from the two commonly used datasets, we test our model on an additional dataset, CelebA, under both unsupervised and semi-supervised settings, which is a more challenging task because attributes of human faces are rather abstract. It turns out that our model still achieves the best results.


We make the following contributions: (1) We are the first to achieve general EM process using GAN by introducing a novel GAN based EM learning framework (GAN-EM) that is able to perform clustering, semi-supervised classification and dimensionality reduction; (2) We conduct thoughtful experiments and show that our GAN-EM achieves state-of-the-art clustering results on MNIST and CelebA datasets, and semi-supervised classification results on SVHN and CelebA. (3) We relax the Gaussian assumption of GMM by applying L-Lipchitz continuity on the generator of GAN.

\section{Related Work}
\label{sec:related}

\textbf{Image Clustering:} Image classification has been well developed due to the advances of Convolutional Neural Network (CNN)~\cite{Alex-Net} in recent years. However, the excellent classification performance relies on large amounts of image labels. Deep models are far from satisfactory in scenarios where the annotations are insufficient. Therefore, image clustering is an essential problem in computer vision studies. \cite{DEC} proposed Deep Embedded Clustering (DEC) which learns feature representations and cluster assignments using deep neural networks. \cite{deep_clustering} is another study on deep clustering, which aims to cluster data into multiple categories by implicitly finding a subspace to fit each class. 

\textbf{Deep EM:} A successful combination of neural networks and EM is the neural expectation maximization~(N-EM)~\cite{Neural-EM}. N-EM trains the parameters of EM using a neural network, which derives a differentiable clustering model, and is used for unsupervised segmentation, where N-EM can cluster constituent objects. Banijamali et al.~\cite{GMN} use generative mixture of networks (GMN) to simulate the GMM. They first use K-means to obtain prior knowledge of the dataset, and then treat each network as a cluster. Variational deep embedding (VaDE)~\cite{VaDE} combines GMM with variational autoencoder (VAE), which keeps the Gaussian assumption. In M-step, VaDE maximizes the lower bound on the log-likelihood given by Jensen inequality. In E-step, a neural network is used to model the mapping from data to class assignment.

\textbf{GAN Clustering:} Generative adversarial networks evolute through the years. At the outset, the vanilla GAN~\cite{GAN} could not perform clustering or semi-supervised classification. Springenberg~\cite{CatGAN} proposed categorical generative adversarial networks~(CatGAN), which can perform unsupervised learning and semi-supervised learning. They try to make the discriminator be certain about the classification of real samples, and be uncertain about that of generated samples, and apply the opposite for the generator. Moreover, their model is based on the assumption that all classes are uniformly distributed, while we relax such assumption in our model. InfoGAN~\cite{InfoGAN}, adversarial autoencoder~\cite{AAE}, feature matching GAN~\cite{improved-tech-for-GAN} and pixelGAN autoencoder~\cite{pixelGAN} are all GAN variants that can do clustering tasks in unsupervised manner. Our proposed GAN-EM is quite different from the previous GAN variants, in which we fit GAN to the EM framework which has been proved that the likelihood of data increases monotonically.
A concurrent work to ours is~\cite{GANMM}. Similar to GMM, they fit GANs into the GMM (GANMM). In GANMM, hard label assignment strategy limits the model to K-means, which is an extreme case of EM for mixture model~\cite{Bishop}. We have three differences from their work. First, we use soft label assignment, rather than the hard assignment in GANMM. To the best of our knowledge, our work is the first to achieve general EM process using GAN. Second, we use only one GAN, rather than $K$ GANs where $K$ is the number of clusters. The drawback of using multiple GANs will be discussed in Sec.~\ref{singleD}. Third, we deal with prior distribution assumption by making the generator L-Lipschitz continuous. Experimental results show that our GAN-EM outperforms GANMM by a big margin.

\section{GAN-EM}
\label{model}

\begin{figure}
\centering
  \includegraphics[width=0.7\textwidth]{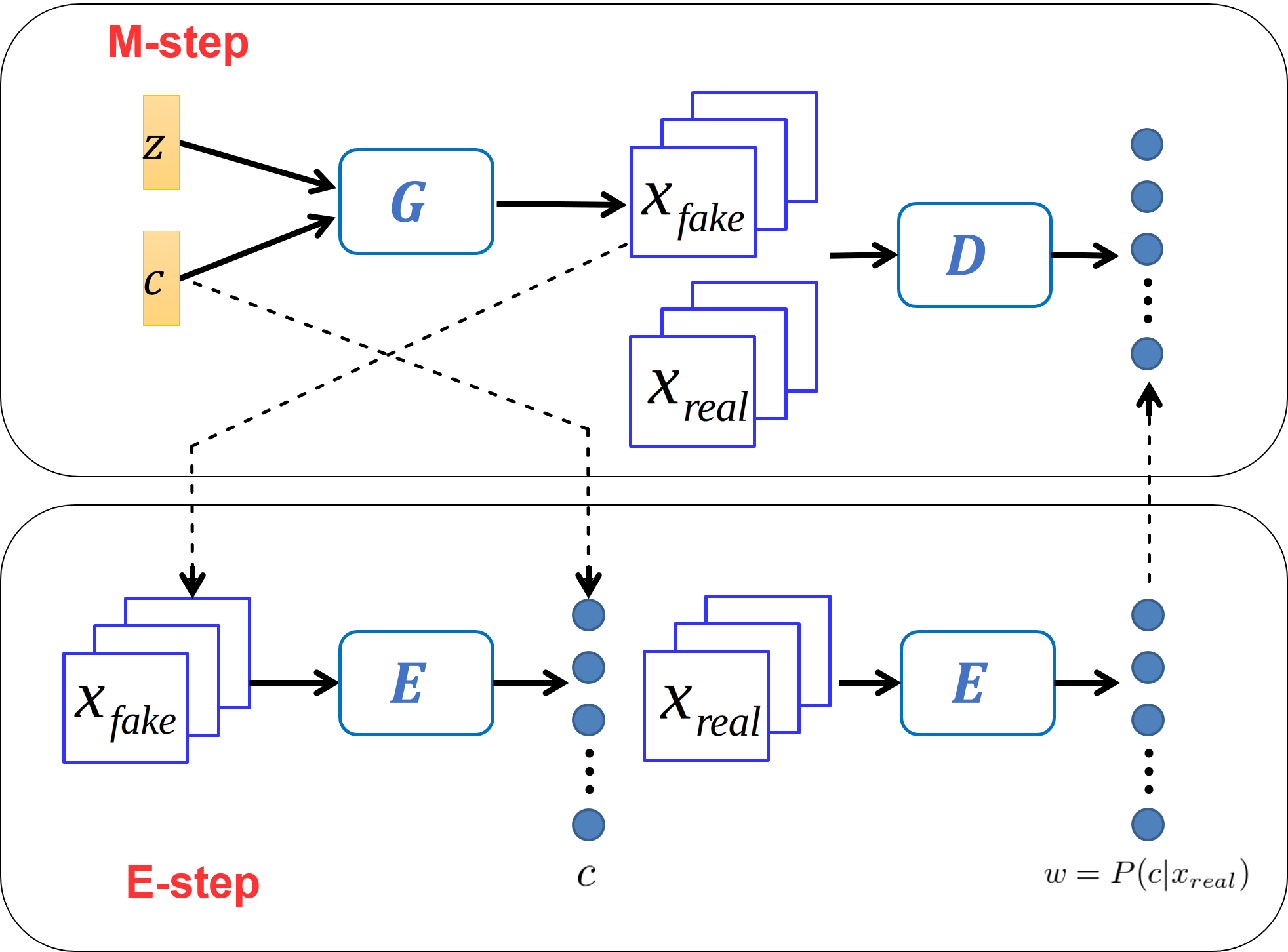}
  \caption{GAN-EM architecture. $G$: generator, $D$: discriminator, $E$: E-net, $z$: random noise, $c$: specified class, $x_{real}$: real images, and $x_{fake}$: generated images. $G$ takes $z$ and $c$ as input and generates $x_{fake}$. The inverse function of $G$ is approximated by $E$, which is trained with input $x_{fake}$ and label $c$. $D$ takes in both $x_{real}$ and $x_{fake}$ and outputs the probability of an input to be real for each class. $E$ is tasked to make soft class assignments to all $x_{real}$.}
  \label{emgan}
\end{figure}

The overall architecture of our model is shown in Fig.~\ref{emgan}. We first clarify some denotations. $\psi$ is the parameters for GAN, which includes $\psi_G$ for generator and $\psi_D$ for discriminator, and $\phi$ is the parameters for multinomial prior distribution, which is composed of
$\phi_i = P(c=i; \phi)$. $\theta=\{\psi, \phi\}$ stands for all the parameters for the EM process. We denote the number of clusters by $K$, and the number of training samples by $N$. 


\subsection{M-step}

The goal of M-step is to update the parameters $\theta$ that maximizes the lower bound of log-likelihood $\log P(x;\theta)$ given the soft class assignment $w=P(c|x;\theta^{old})$ provided by E-step (where $w$ is a $N\times K$ matrix). $\theta$ consists of $\phi$ and $\psi$. Updating $\phi$ is simple, since we can compute the analytical solutions for each $\phi_i$: $\phi_i^* = N_i/N$, where $N_i$ is the number of samples for the $i$-th cluster. Details are in Sec.~\ref{sec:theory}. 

To update $\psi$, we extend GAN to a multi-class version to learn the mixture models $P(x|c=i; \psi)\ for\ i=1\dots K$. The vanilla GAN~\cite{GAN} is modified in several ways as follows.

\noindent \textbf{Generator:} \quad
Similar to conditional GAN~\cite{CGAN}, apart from the random noise $z$,  class label information $c$ is also added to the input of the generator. With $c$ as input, we specify the generator to generate the $c$-th class images. This makes our generator act as $K$ generators.

\noindent \textbf{Discriminator:} \quad 
Different from the vanilla GAN that uses only one output unit, we set $K$ units in the output layer.

\noindent \textbf{Loss function:} \quad 
Here, we only give the form of the loss functions, the derivation of which will be discussed in Sec.~\ref{em theory} in detail. $L_G$ and $L_D$ are loss functions of the generator and the discriminator respectively. We have: 
\begin{align}
L_G = &-\frac{1}{2}\mathop{\mathbb{E}} \limits_{c\sim U}\mathop{\mathbb{E}} \limits_{z\sim \mathcal{N}}\exp\{\sigma^{-1}[D_c(G(z,c))]\} \enspace, \label{gloss}\\ 
L_D = &\mathop{\mathbb{E}} \limits_{c\sim U}\mathop{\mathbb{E}} \limits_{z\sim \mathcal{N}}\sum_{i=1}^{K}\log(1-D_i(G(z, c))) \nonumber \\
&+ \mathop{\mathbb{E}} \limits_{x\sim P_{r}}\sum_{i=1}^{K}  w_i \log(D_i(x))
\enspace,
\label{dloss}
\end{align}
where $z$ is random noise, $c$ is  the specified class of images to be generated, $\textit{U}$ and $\mathcal{N}$ are uniform distribution and Gaussian distribution respectively, $x \sim P_r$ is to sample images from the real data distribution. $G(z,c)$ stands for generated images, $\sigma$ for sigmoid activation function, and $w_i=P(c=i|x;\theta^{old})$ for the $i$-th class label assignments, which comes from the previous E-step. Here, $D_i(\cdot)$ denotes the $i$-th output unit's value of the discriminator. Notice that $L_G$ is a modified form for the loss of the generator so that GAN can perform likelihood maximization~\cite{gantutorial}. 
The first term of $L_D$ means that all output units are expected to give a low probability to fake images. Conversely, the second term guides the discriminator to output a high probability for all real ones, while the loss for each output unit is weighted by the soft label assignment $w$ passed from E-step. 
\subsection{E-step}
In the unsupervised manner, class label $c$ for real data cannot be observed and is regarded as latent variable. Thus the goal of E-step is to estimate such latent variable, which is normally obtained using Bayes rule given the parameters learned from the previous M-step. Therefore, we have: 
\begin{align}
P(c=i|x; \theta)=\frac{P(x|c=i; \psi)P(c=i;\phi)}{\sum_{j}P(x|c=j;\psi)P(c=j;\phi)}
\enspace,
\label{e-step}
\end{align}
where $P(x|c=i;\psi)$ represents the distribution of data in class $i$ given by the generator network, and the denominator is the normalization term. However, Eq.~\ref{e-step} is hard to calculate since neural network is not invertible. 

To circumvent such problem, we introduce another neural network, called E-net, to fit the distribution expressed on the left hand side of Eq.~\ref{e-step}. To train the E-net, we first generate samples from the generator, where the number of generated samples for cluster $i$ is subject to $\phi_i$ because $P(c=i|x; \theta)$ is proportional to $\phi_i$ according to Eq.~\ref{e-step} (remind that $\phi_i=P(c=i;\phi)$). After the E-net is well trained, it approximates the parameters $\theta$ and act as an inverse generator. Similar approach is also used in BiGAN~\cite{Advesarial-feature-learning}, where they also prove that $E=G^{-1}$ almost everywhere. However, the goal of BiGAN is feature learning, which is different from ours.

Specifically, as shown in Fig.~\ref{emgan} we take the output of the generator, i.e., $x_{fake}=G(z,c)$, as the input of the E-net, and take the corresponding class $c$ as the output label.
Therefore, we can learn the approximate distribution of the left hand side of Eq.~\ref{e-step}, and thus obtain soft class assignments:
\begin{align}
w=P(c|x_{real}; \theta)
\enspace,
\end{align}
then feed them back to M-step. The E-net takes the following loss:
\begin{align}
L_{E}= \mathop{\mathbb{E}} \limits_{z\sim \mathcal{N}}\mathop{\mathbb{E}} \limits_{c\sim \mathcal{\phi}}\text{CE}\{E(G(z,c)), u\} \enspace,
\end{align}
where $\text{CE}\{\cdot, \cdot\}$ stands for cross-entropy function, $u$ is a one-hot vector that encodes the class information $c$, and $E(\cdot)$ is the output of E-net. The trained E-net is then responsible for giving the soft class assignment $w$ for real images. 

\subsection{EM algorithm}

The bridge between M-step and E-step is the generated fake images with their conditional class labels and the output $w$ produced by E-net. So far, the whole training loop has been built up. Then we can train M-step and E-step alternatively to achieve the EM mechanism without Gaussian assumption. We start the training with M-step, where $w$ is initialized with uniform distribution. The pseudo code is in Algorithm~\ref{alg:emgan}.

\begin{algorithm}
\caption{GAN-EM}
\label{alg:emgan}
\begin{algorithmic}[1]
    \State \textbf{Initialization:} $w_i=1/K\ for\ i=1\dots K$
    \For{$iteration=1\dots n$}
    	\State\textbf{update $\phi$:}\Comment{M-step}
        \State\hspace{\algorithmicindent}$\phi_i = N_i/N\ for\ i=1\dots K$
        \State\textbf{update $\psi$:}
    	\State\hspace{\algorithmicindent}\textbf{for}\ $epoch = 1\dots p$\ \textbf{do}
        	\State\hspace{\algorithmicindent}\hspace{\algorithmicindent}train G: $\min\limits_{\psi_G}L_G$
            \State \hspace{\algorithmicindent}\hspace{\algorithmicindent}train D: $\min\limits_{\psi_D}L_D$
      	\State\hspace{\algorithmicindent}\textbf{end for}
        \For{$epoch = 1\dots q$}\Comment{E-step}
            \State Sample a batch of $c\sim \phi$, and obtain $G(z,c)$
            \State train E-net: $\min\limits_{\eta} L_E$ ($\eta$: weights of E-net\footnotemark)
        \EndFor
        \State \textbf{update label assignment:} $w = E(x_{real})$
    \EndFor
    \State \textbf{Predict:} $w = E(x_{real})$
\end{algorithmic}
\end{algorithm}

\section{Theoretical Analysis}
\label{sec:theory}
This section mainly focuses on the theoretical analysis of GAN in M-step. We first show how our model works with $K$ GANs by deriving the objective functions. After that, we simplify the model by using only one GAN. Finally, we show how to deal with prior distribution assumption.
\footnotetext{Because E-net aims to learn an inverse function of generator, $\eta$ is not independent with $\psi$. Thus, we could not say that $\eta$ is the parameter of EM process. In other words, $\eta$ is not part of $\theta$.}
\subsection{Background}
\label{em theory}

In M-step, we aim to maximize the Q function~\cite{Bishop} expressed as: 
\begin{align}
&Q(\theta;\theta^{old}) = \mathop{\mathbb{E}} \limits_{x\sim P_r} \sum_{i=1}^{K} P(c=i|x;\theta^{old}) \log P(x, c=i;\theta) \nonumber\\
&=\mathop{\mathbb{E}} \limits_{x\sim P_r} \sum_{i=1}^{K}w_i\log [P(x|c=i;\psi)P(c=i;\phi)]
\enspace.
\label{qfunction}
\end{align}
Furthermore, we can write Eq.~\ref{qfunction} as the sum of two terms $Q^{(1)}$ and $Q^{(2)}$:
\begin{align}
&Q^{(1)} = \mathop{\mathbb{E}} \limits_{x\sim P_r} \sum_{i=1}^{K}w_i\log P(x|c=i;\psi) \enspace, \label{Q1} \\
&Q^{(2)} = \mathop{\mathbb{E}} \limits_{x\sim P_r} \sum_{i=1}^{K}w_i\log \phi_i \enspace. \label{Q2}
\end{align}
Remind that $\phi_i = P(c=i;\phi)$ is explained in the previous section. These two terms are independent with each other, so they can be optimized separately. 

We first optimize Eq.~\ref{Q2}. Since $\phi_i$ is irrelevant to $x$, we can ignore the expectation because it only introduces a constant factor. With the constraint $\sum_i \phi_i = 1$, the optimal solution for $Q^{(2)}$ is $\phi_i^*=N_i/N\ (i=1\dots K)$, where $N_i$ is the sample number of the $i$-th cluster ($N_i$ is not an integer necessarily because the class label assignment is in a soft version), and $N$ is the sample number of the whole dataset. Derivation can be seen in Appendix~\ref{phi opt proof}.

Then we consider Eq.~\ref{Q1}. We are expecting to employ GANs to optimize $Q^{(1)}$. Each term in the summation in $Q^{(1)}$ is independent with each other because we are currently considering $K$ separate GANs. Therefore, we can optimize each term in $Q^{(1)}$, rewrite it as
\begin{align}
Q_i^{(1)} = \mathop{\mathbb{E}} \limits_{x\sim P_r} w_i\log P_{f_i}(x|c = i; \psi_i)
\enspace,
\label{Q1-func-each-term}
\end{align}
to fit the single GAN model, and sum them up in the end. Here, $\psi_i$ is the parameters of the $i$-th GAN, $P_r$ stands for the distribution of all real data, and $P_{f_i}$ stands for the fake data distribution for each cluster $c = i$.

For convenience, we introduce a new distribution 
\begin{align}
P_{r_i} = \frac{1}{Z} w_iP_r
\enspace,
\label{pri}
\end{align}
where $\frac{1}{Z}$ is the normalization factor, $Z = \int_x w_iP_r dx$. Substitute $P_{r_i}$ into Eq.~\ref{Q1-func-each-term} and we have: 
\begin{align}
Q_i^{(1)} = Z \mathop{\mathbb{E}} \limits_{x\sim P_{r_i}}\log P_{f_i}(x|c = i; \psi_i)
\enspace.
\label{glanz-before}
\end{align}
We can drop the constant $Z$ to obtain: 
\begin{align}
\widetilde{Q}_i^{(1)} = \mathop{\mathbb{E}} \limits_{x\sim P_{r_i}}\log P_{f_i}(x|c = i; \psi_i)
\enspace, 
\label{glanz-distribution}
\end{align}
which is equivalent to Eq.~\ref{glanz-before} in terms of optimization.
\subsection{Objectives}
\label{objectives}
In this subsection, we aim to show how our design for M-step (i.e. Eq.~\ref{gloss} and Eq.~\ref{dloss}) is capable of maximizing Eq.~\ref{glanz-distribution} which takes exactly the form of likelihood. This is feasible due to the following two facts~\cite{gantutorial}: 
\begin{enumerate}
\item MLE is equivalent to minimizing the KL-divergence between the real distribution and generated distribution;
\item When discriminator is optimal, we can modify the loss function for generator so that the optimization goal of GAN is to minimize KL divergence.
\end{enumerate}
Maximizing Eq.~\ref{glanz-distribution} is equivalent to minimizing $KL(P_{r_i}\parallel P_{f_i})$ according to fact 1. Then we show that how GANs can be tasked to minimize such KL-divergence by introducing minor changes.

%
According to fact 2, only when discriminator is optimized can we modify GAN to minimize KL divergence. Therefore, we consider the optimum of discriminators. With Eq.~\ref{pri}, the loss function of the $i$-th discriminator given any generator (denoted by fake data) is:
\begin{align}
L_i &= \mathop{\mathbb{E}} \limits_{x\sim P_{r}} w_i \log D_i(x) + \mathop{\mathbb{E}} \limits_{x\sim P_{f_i}} \log (1-D_i(x)) \nonumber \\
&= \mathop{\mathbb{E}} \limits_{x\sim P_{r_i}} \log D_i(x) + \mathop{\mathbb{E}} \limits_{x\sim P_{f_i}} \log (1-D_i(x))
\enspace,
\label{Loss-ith-discriminator}
\end{align}
where $D_i(\cdot)$ is the $i$-th discriminator, similar to vanilla GAN~\cite{GAN}. We show that when the discriminators reach optimum, $L$ is equivalent to the sum of JS divergence. The following corollary is derived from Propositional 1 and Theorem 1 in~\cite{GAN}. 


\begin{corollary} 
\label{thm1}
Equation~\ref{Loss-ith-discriminator} is equivalent to the JS divergence between real distribution and generated distribution for each cluster when discriminators are optimal, i.e.
\begin{align}
L_i=-(w_i + 1)\log 2 + 2JSD(P_{r_i}\parallel P_{f_i})
\enspace,
\label{jsd}
\end{align}
and the optimal $D_i^*$ for each cluster is: 
\begin{align}
D_i^*(x) = \frac{P_{r_i}(x)}{P_{r_i}(x) + P_{f_i}(x)}
\enspace.
\label{D-opt}
\end{align}
\end{corollary}

\begin{proof}
See Appendix~\ref{thm1 proof}.
\end{proof}

If we use the same loss function for generator as the vanilla GAN, the JS divergence in Eq.~\ref{jsd} will be minimized. However, we aim to make GAN to minimize the KL divergence, $KL(P_{r_i}\parallel P_{f_i})$, for each cluster so as to achieve the goal of maximizing Eq.~\ref{glanz-distribution}. In Corollary.~\ref{thm1}, we already have the optimal discriminator given fixed generator. Therefore, according to fact 2, we need to modify the loss function for the generator as: 
\begin{align}
-\frac{1}{2}\mathbb{E}_z\exp\{\sigma^{-1}[D_i(G_i(z))]\}
\enspace,
\label{mle-gan}
\end{align}
where $\sigma$ is the sigmoid activation function in the last layer of the discriminator, and $G_i(\cdot)$ is the output of $i$-th generator.

Now we have derived the objectives of single generator and discriminator, and we need to ensemble them up as a whole model. Since we are currently using $K$ GANs, we only need to sum up Eq.\ref{mle-gan} for the loss of generators:
\begin{align}
-\frac{1}{2}\sum_{i=1}^{K}\mathbb{E}_z\exp\{\sigma^{-1}[D_i(G_i(z)]\} \enspace,
\label{sum-mle-gan}
\end{align}
and sum up Eq.~\ref{Loss-ith-discriminator} for the loss of discriminators:
\begin{align}
\sum_{i=1}^{K} \mathop{\mathbb{E}} \limits_{x\sim P_{r}} w_i \log D_i(x) + \mathop{\mathbb{E}} \limits_{x\sim P_{f_i}} \log (1-D_i(x))
\enspace.
\label{kdiscriminator}
\end{align}

Here, Eq.~\ref{kdiscriminator} is equivalent to Eq.~\ref{dloss} since $x\sim P_{f_i}$ is generated by generator $G$. The derivation from Eq.~\ref{sum-mle-gan} to Eq.~\ref{gloss} will be introduced in Sec.~\ref{singleD}.

\subsection{Single GAN v.s. multiple GANs}
\label{singleD}

We have shown that $K$ GANs can be tasked to perform MLE in M-step of EM. In fact, using such many GANs is intractable since the complexity of the model grows along with cluster numbers proportionally. Moreover, data is separated per cluster and distributed to different GANs, which could not make the most use of data for individual GAN efficiently. 

\subsubsection{Single generator}
For the generator part, we employ a conditional variable $c$~\cite{CGAN} to make a single generator act as $K$ generators. Then the final loss function for generator is exactly Eq.~\ref{gloss}.

\subsubsection{Single discriminator}
In our work, instead of applying $K$ discriminators, we use a single discriminator with $K$ output units. Each output has individual weights in the last fully connected layer. The preceding layers of the discriminator is shared since the convolutional layers play a role in extracting features that are often in common among different clusters of data.

To this end, we denote the last fully connected layer of the discriminator by function $\tilde{D}$, and all other layers by function $f$, then we have: 
\begin{align}
D_i(x) = \widetilde{D}_i(f(x)) = \widetilde{D}_i(\tilde{x})
\enspace, \enspace \forall i\in \{1,\dots,K\} \enspace, 
\end{align}
where $\tilde{x} = f(x)$ stands for the features learned by $f$. The objective still holds the form of Eq.~\ref{kdiscriminator}, but the meaning of $D_i$ has changed from the $i$-th discriminator to the $i$-th output unit of $D(x)$ that stands for the probability of $x$ belonging to $i$-th cluster. Till now, we have finished deriving the loss functions for our proposed model.

In practice, to speed up the learning, we add an extra output unit for the discriminator, which only distinguishes all fake data from all real data.

\subsection{Prior distribution assumption}

The prior distribution assumption is necessary because without one we will have $\theta^* = \theta^{old}$ for the MLE in M-step and then the parameters of the model will not be updated anymore. GMM has a strong assumption on prior distribution, i.e. Gaussian distribution, while our GAN-EM has a weaker assumption. Details are discussed as follows.

In our work, we make a conjecture that the data distribution generated by a well-trained GAN obeys intra-cluster continuity and inter-cluster discontinuity. Such assumed property of GAN can be seen as the prior distribution assumption of GAN-EM, which is exactly the goal of clustering. Also, compared to GMM, GAN is more powerful in terms of modeling distribution. Therefore, by applying GAN to the EM process, we weaken the Gaussian assumption of GMM. Next, we discuss the intuition of the conjecture.

For intra-cluster continuity, we make use of the uniform continuity of generator. In fact, we use CNN for generator where each layer can be treated as a uniformly continuous function. Composition of these functions, i.e. the generator, is then also uniformly continuous. Therefore, the output space of generator is continuous since the input Gaussian noise is continuous.

For inter-cluster discontinuity, the discriminator helps prevent generator from disobeying this rule. For convenience, we call fake data lying in the gap between clusters ``gap data''. Suppose that the generator disobeys inter-cluster discontinuity, e.g. the generator treats two clusters as one. Then it must generate gap data to maintain the intra-cluster continuity. In this case, a well trained discriminator penalizes the generator harder because gap data are farther away from real data, so that the generator will eventually obey inter-cluster discontinuity.

However, in practice, we may encounter a situation where the generator generates very sparse gap data which also satisfies intra-cluster continuity, but the penalization given by discriminator is too small due to the sparsity. Consequently, the clustering would be led to a wrong direction. To solve the problem, we can make the generator L-Lipschitz continuous, which is much stronger than uniformly continuous. We use weight clipping to enforce Lipschitz condition similar to what WGAN~\cite{WGAN} does.

\section{Experiments}

We perform unsupervised clustering on MNIST~\cite{MNIST} and CelebA~\cite{celeba} datasets, and semi-supervised classification on MNIST, SVHN~\cite{svhn} and CelebA datasets. We also evaluate the capability of dimensionality reduction by adding an additional hidden layer to the E-net. The results show that our model achieve state-of-the-art results on various tasks. Meanwhile, the quality of generated images are also comparable to many other generative models. The training details and network structures are illustrated in Appendix~\ref{imp-detail}.

\subsection{Implementation details}
We apply RMSprop optimizer to all 3 networks G, D and E with learning rate 0.0002 (decay rate: 0.98). The random noise of generator is in uniform distribution. In each M-step, there are 5 epoches with a minibatch size of 64 for both the generated batch and the real samples batch. We use a same update frequency for generator and discriminator. For E-step, we generate samples using well trained generator with batch size of 256, then we apply 1000 iterations to update E-net.
\subsection{Unsupervised clustering}

GAN-EM achieves state-of-the-art results on MNIST clustering task with 10, 20 and 30 clusters. We evaluate the error rate based on the following metric which has been used in most other clustering models in Tab.~\ref{result}:
\begin{align}
Err = 1 - \max \limits_{m \in \mathcal{M}}\frac{\sum_{i=1}^{K}\mathds{1}\{l_{i} = m(c_i)\}}{K}
\enspace,
\end{align}
where $l_{i}$ is for the predicted label of the $i$-th cluster, $c_i$ for ground truth label, and $\mathcal{M}$ for all one-to-one mapping from ground truth labels to predicted labels. We compare our method with most popular clustering models that use generative models like GANs or autoencoders. The experimental results are shown in column 1 of Tab.~\ref{result}. 

\begin{table*}
\centering
\caption{Experiment results of different models on MNIST and SVHN.}
\label{result}
\begin{tabular}{l|lll|l}
\hline
                                                                & \begin{tabular}[c]{@{}l@{}}MNIST\\ (Unsupervised)\end{tabular} & \begin{tabular}[c]{@{}l@{}}MNIST\\ (100 labels)\end{tabular} & \begin{tabular}[c]{@{}l@{}}MNIST\\ (1000 labels)\end{tabular} & \begin{tabular}[c]{@{}l@{}}SVHN\\ (1000 labels)\end{tabular} \\ \hline
K-means~\cite{pixelGAN}                 & 46.51 ($K=10$)                                       & -                                              & -                                                             & -                                              \\
GMM~\cite{pixelGAN}                 & 32.61($\pm$0.06) ($K=10$)                                       & -                                             & -                                                             & -                                              \\
DEC~\cite{DEC}                 & 15.7 ($K=10$)                                       & -                                              & -                                                             & -                                              \\

VAE~\cite{VAE}                           & -                                                              & 3.33($\pm$0.14)                                              & 2.40($\pm$0.02)                                               & 36.02($\pm$0.10)                                             \\
AAE~\cite{AAE}                            & 4.10($\pm$ 1.13) ($K=30$)                                      & 1.90($\pm$0.10)                                              & 1.60($\pm$0.08)                                               & 17.70($\pm$0.30)                                             \\
CatGAN~\cite{CatGAN}                      & 4.27 ($K=20$)                                                  & 1.91($\pm$0.10)                                              & 1.73($\pm$0.28)                                               & -                                                            \\
InfoGAN~\cite{InfoGAN}                   & 5.00 (not specified)                                           & -                                                            & -                                                             & -                                                            \\
Improved GAN~\cite{improved-tech-for-GAN} & -                                                              & \textbf{0.93($\pm$0.06)}                                              & -                                                             & 8.11($\pm$1.30)                                              \\
VaDE~\cite{VaDE}                          & 5.54 ($K=10$)                                                  & -                                                            & -                                                             & -                                                            \\
PixelGAN~\cite{pixelGAN}                 & 5.27($\pm$1.81) ($K=30$)                                       & 1.08($\pm$0.15)                                              & -                                                             & 6.96($\pm$0.55)                                              \\
GANMM~\cite{GANMM}                       & 35.70($\pm$0.45) ($K=10$)                                      & -                                                            & -                                                             & -                                                            \\ \hline
\textbf{GAN-EM}                                & 4.20($\pm$ 0.51) ($K=10$)                                      & 1.09($\pm$ 0.18)                                             & \textbf{1.03($\pm$ 0.15)}                                              & \textbf{6.05($\pm$ 0.26)}                                             \\
                                                                & \textbf{4.04($\pm$ 0.42)} ($K=20$)                                      & -                                                            &                                                               & -                                                            \\
                                                                & \textbf{3.97($\pm$ 0.37)} ($K=30$)                                      & -                                                            &                                                               & -                                                            \\ \hline
\end{tabular}
\end{table*}

\begin{figure}
\begin{subfigure}{0.48\textwidth}
\includegraphics[width=\textwidth]{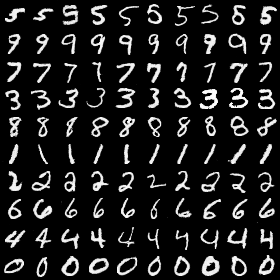}
\caption{MNIST (unsupervised)}\label{mnist}
\end{subfigure}
\hspace*{\fill} 
\begin{subfigure}{0.48\textwidth}
\includegraphics[width=\textwidth]{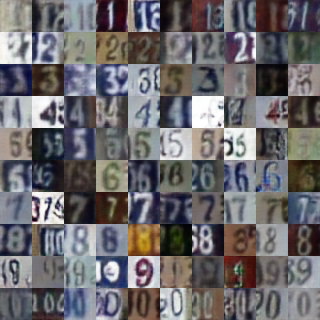}
\caption{SVHN (1000 labels)}\label{svhna}
\end{subfigure}
\hspace*{\fill} 

\caption{Clustering and semi-supervised classification results by GAN-EM.} \label{show}
\end{figure}

\begin{table}
\centering
\caption{GAN-EM unsupervised clustering and semi-supervised classification on all 40 CelebA attributes}
\begin{tabular}{l|l|l}
\hline
         & unsupervised & \begin{tabular}[c]{@{}l@{}}semi-supervised\\ (100 labels)\end{tabular} \\ \hline
VAE      & -            & 45.38                                                                  \\
AAE      & 42.88        & 31.03                                                                  \\
CatGAN   & 44.57        & 34.78                                                                  \\
VaDE     & 43.64        & -                                                                      \\
PixelGAN & 44.27        & 32.54                                                                  \\
GANMM    & 49.32        & -                                                                  \\ \hline
\textbf{GAN-EM}    & \textbf{42.09}        & \textbf{28.82}                                                                  \\ \hline
\end{tabular}
\label{CelebA}
\end{table}

Since different models have different experiment setups, there is no uniform standard for the clustering numbers under the unsupervised setting. MNIST dataset has 10 different digits, so naturally we should set the number of clusters $K=10$. However, some models such as CatGAN, AAE, and PixelGAN use $K=20$ or $30$ to achieve better performance, since the models might be confused by different handwriting styles of digits. In other words, the more clusters we use, the better performance we can expect. To make fair comparisons, we conduct experiments with $K=10, 20, 30$ respectively. Also, all models in Tab.~\ref{result} take input on the 784-dimension raw pixel space. Note that both K-means and GMM have high error rates (46.51 and 32.61) on raw pixel space, since MNIST is highly non-Gaussian, which is an approximate Bernoulli distribution with high peaks at 0 and 1. The huge margin achieved by GAN-EM demonstrates that the relaxation of Gaussian assumption is effective on clustering problem.

Our proposed GAN-EM has the lowest error rate 3.97 with $K=30$. Moreover, When $K=20$, GAN-EM still has better results than other models. With 10 clusters, GAN-EM is only outperformed by AAE, but AAE achieves the error rate of 4.10 using 30 clusters. VaDE also has a low error rate under the setting of $K=10$, yet still higher than that of our model under the same setting. GANMM has a rather high error rate\footnote{When the feature space is reduced to 10 dimensions using SAE, GANMM achieves an error rate of 10.92 ($\pm 0.15$) with $K=10$~\cite{GANMM}.}, while GAN-EM achieves the state-of-the-art clustering results on MNIST, which shows that the clustering capability of EM is much superior to that of K-means.

\begin{figure*}
\begin{subfigure}{\textwidth}
\includegraphics[width=\textwidth]{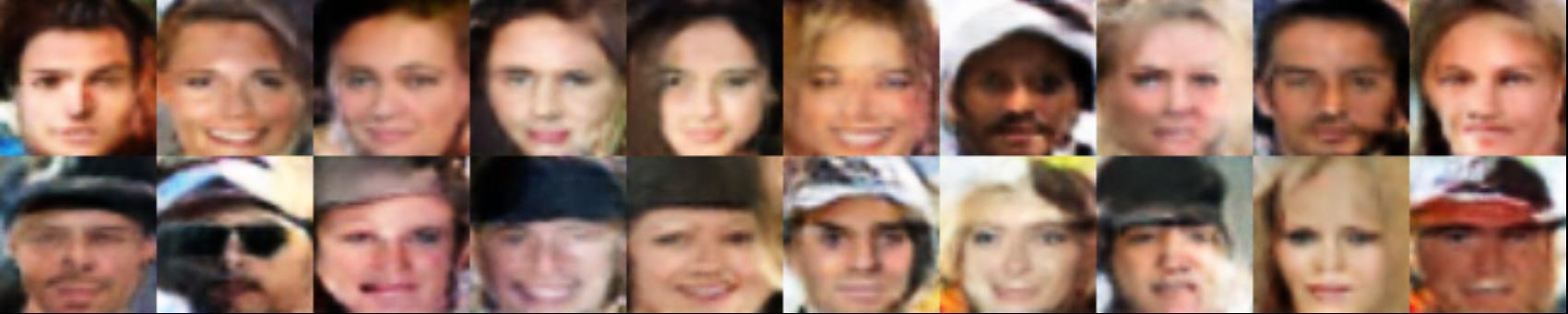}
\caption{CelebA hat}\label{celebaa}
\end{subfigure}
\begin{subfigure}{\textwidth}

\includegraphics[width=\textwidth]{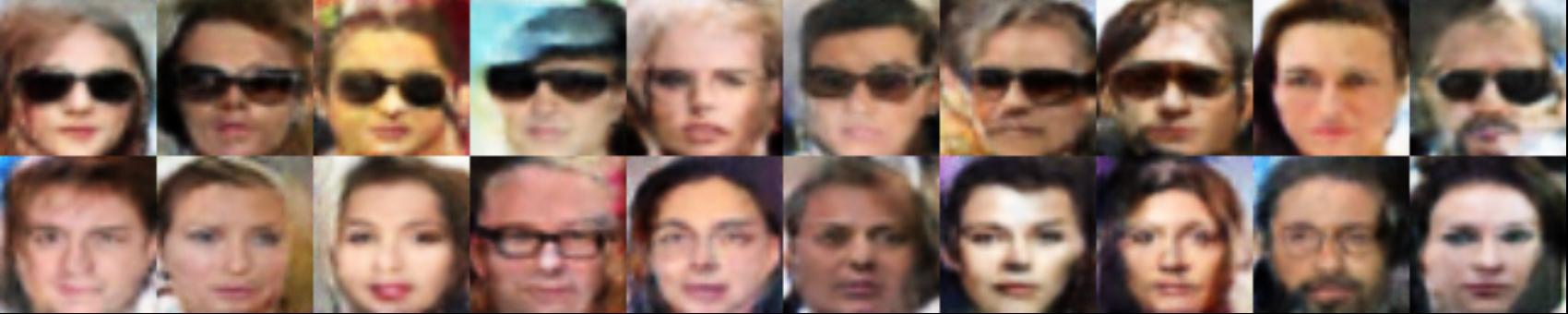}
\caption{CelebA glasses}\label{celebab}
\end{subfigure}
\caption{Unsupervised feature learning on CelebA: (a) generated images after clustering learning on 'hat' attribute; 
(b) generated images after clustering learning on 'glasses' attribute.} \label{celeba}
\end{figure*}

Then we test our model on CelebA dataset using the same strategy as stated above. CelebA is a large-scale human face dataset that labels faces by 40 binary attributes. Totally unsupervised clustering on such tasks is rather challenging because these face attributes are so abstract that it is difficult for CNN to figure out what features it should extract to cluster the samples. Tab.~\ref{CelebA} (column 1) illustrates the average error rate of different models on all 40 CelebA attributes. We also list detailed results of GAN-EM on all the 40 attributes in Appendix~\ref{emgan-celeba-results}.
We achieve the best overall result for unsupervised clustering, and we demonstrate two representative attributes on which we achieve lowest error rates, i.e. hat (29.41) and glasses (29.89), in Figs.~\ref{celebaa} and~\ref{celebab}, where the two rows of images are generated by the generator given two different conditional labels respectively. The details of strategies for selecting samples is illustrated in supplementary material Appendix~\ref{celeba-appendix}.

\subsection{Semi-supervised classification}

It is easy to extend our model to semi-supervised classification tasks where only a small part of samples' labels are known, while the remainders are unknown. We use almost the same training strategies as clustering task except that we add the supervision to the E-net in every E-step. The method is that at the end of the E-net training using generated fake samples, we train it by labeled real samples. Then the loss function takes the form $L_{E}= CE\{E(x_{real}), u\}$, where $u$ is the one-hot vector that encodes the class label $c$. Once the E-net has an error rate below $\epsilon$ on the labeled data, we stop the training, where $\epsilon$ is a number that is close to zero (e.g. 5\% or 10\%) and can be tuned in the training process. The reason why $\epsilon$ is greater than zero is to avoid over-fitting. 

We evaluate the performance of semi-supervised GAN-EM on MNIST, SVHN and CelebA datasets. As shown in Tab.~\ref{result} (column 2, 3 and 4) and Tab.~\ref{CelebA} (column 2), our GAN-EM achieves rather competitive results on semi-supervised learning on all three datasets (state-of-the-art on SVHN and CelebA). The images generated by GAN-EM on SVHN are shown in Fig.~\ref{svhna}.
On MNIST, when we use 100 ground truth labels for the semi-supervised classification, the error rate is 1.09, which is only 0.16 higher than the top-ranking result by improved GAN, and when 1000 ground truth labels are used, GAN-EM achieves the lowest error rate 1.03. On SVHN dataset, 1000 labels are applied as other models do, and we achieve state-of-the-art result with an error rate of 6.05.
For CelebA, the number of ground truth labels is set to 100, and our model outperforms all other models with respect to average error rate on all 40 attributes.

\begin{figure}[h!]
\centering
\begin{subfigure}{0.23\textwidth}
\centering
\includegraphics[width=\textwidth]{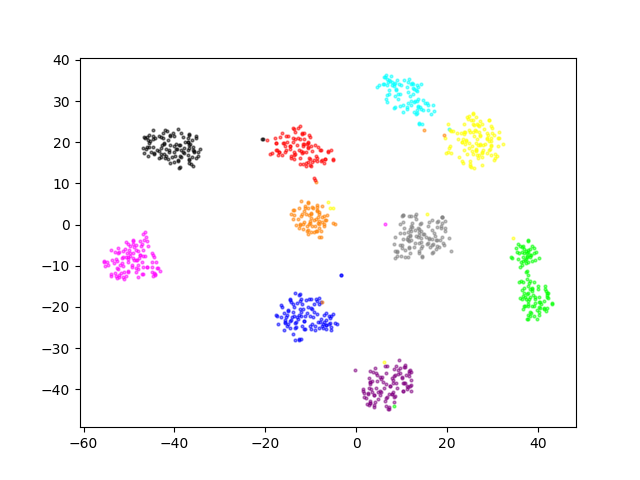}
\caption{Supervised 1000D}\label{tsnia}
\end{subfigure}
\hspace*{\fill} 
\begin{subfigure}{0.23\textwidth}
\centering
\includegraphics[width=\textwidth]{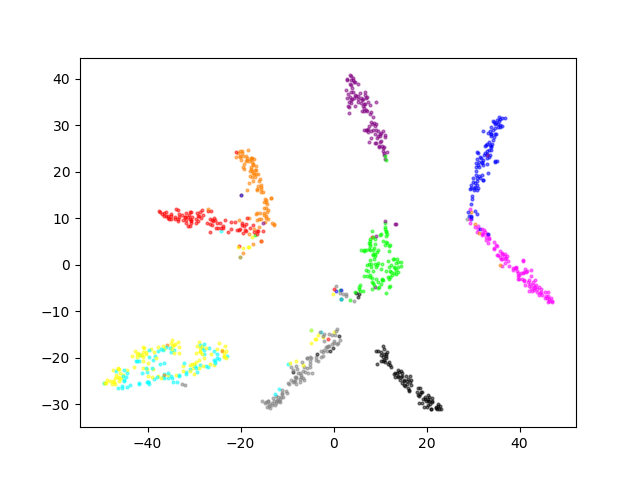}
\caption{Unsupervised 2D}\label{tsnib}
\end{subfigure}
\begin{subfigure}{0.23\textwidth}
\centering
\includegraphics[width=\textwidth]{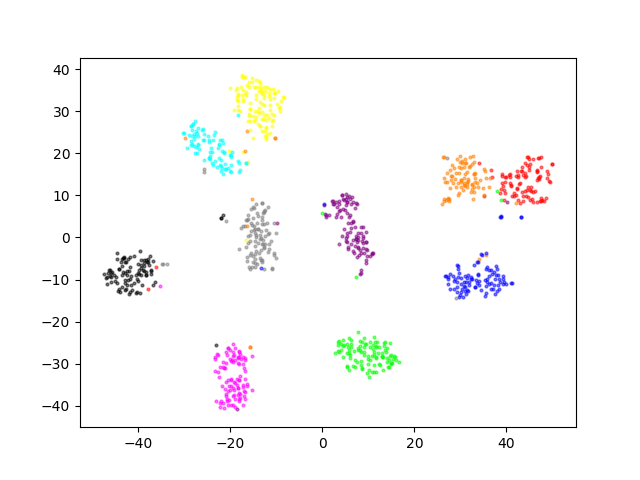}
\caption{Unsupervised 100D}\label{tsnic}
\end{subfigure}
\hspace*{\fill} 
\begin{subfigure}{0.23\textwidth}
\centering
\includegraphics[width=\textwidth]{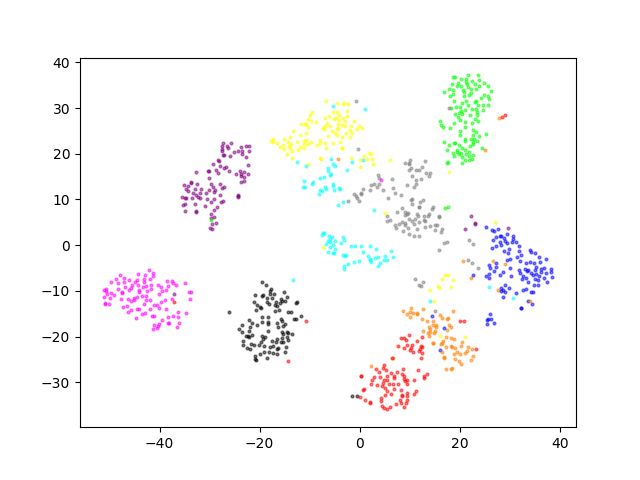}
\caption{Unsupervised 1000D}\label{tsnid}
\end{subfigure}
\caption{Representation of unsupervised dimensionality reduction on MNIST. Each color denotes one class of digit.} \label{tsni}
\end{figure}
\subsection{Dimensionality reduction}


We can easily modify our GAN-EM model to perform dimensionality reduction by inserting a new layer $r$ with $k$ hidden units to the E-net ($k$ is the number of dimension that we want to transform to). Layer $r$ lays right before the output layer. Then, we can use exactly the same training strategy as the unsupervised clustering task. Once the training converges, we consider the E-net as a feature extractor by removing the output layer. Then we feed the real samples to the E-net and take the output on layer $r$ as the extracted features after dimensionality reduction. Three different dimensions, i.e. 1000, 100 and 2, are selected for test on the MNIST dataset. We also apply t-SNE~\cite{tsne} technique to project the dimensionality reduced data to 2D for visualization purpose.

Fig.~\ref{tsnia} shows the supervised feature learning result. Figs.~\ref{tsnib},~\ref{tsnic} and~\ref{tsnid} are unsupervised data dimensionality reduction results with three different dimensions. We can see that our model can deal with all cases very well. Most different digits have large gap with each other and the same digits are clustered together compactly. In particular, the 100 dimension result (Fig.~\ref{tsnic}) has almost equivalent performance with the supervised feature learning (Fig.~\ref{tsnia}). It is worth mentioning that in the 2 dimension reduction case, the second last hidden layer of the E-net has only 2 hidden units, but the clustering error rate is as low as 11.8 with 10 clusters that demonstrates the robustness of GAN-EM model.

\section{Conclusion}

In this paper, we propose a novel GAN-EM learning framework that embeds GAN into EM algorithm to do clustering, semi-supervised classification and dimensionality reduction. We achieve state-of-the-art results on MNIST, SVHN and CelebA datasets. Furthermore, the fidelity of the generated images are comparable to many other generative models. Although all our experiments are performed based on vanilla GAN, GAN-EM framework can also be embedded by many other GAN variants and better results are expected.

\bibliographystyle{plain}  
\bibliography{bibtex}
\newpage

\begin{appendices}
\section{Missing details}
\subsection{Proof for Corollary~\ref{thm1}}
\label{thm1 proof}
\begin{proof} For every cluster $i$, we have:
\begin{align*}
L_i &= \int_{x}w_i P_r(x) \log D_i(x) + P_{f_i}(x) \log (1-D_i(x))dx\\
&= \int_{x}P_{r_i}(x) \log D_i(x) + P_{f_i}(x) \log (1-D_i(x))dx
\enspace.
\end{align*}
Let $\frac{\partial L_i}{\partial D_i} = 0$, we have: 
\begin{align*}
D_i^*(x) = \frac{P_{r_i}(x)}{P_{r_i}(x) + P_{f_i}(x)}.
\end{align*}
Then we substitute this optimal into the summation in Eq.~\ref{Loss-ith-discriminator}:
\begin{align*}
L =&\int_{x} P_{r_i}(x) \log \frac{P_{r_i}(x)}{P_{r_i}(x) + P_{f_i}(x)}\\ 
&+ P_{f_i}(x) \log \frac{P_{f_i}(x)}{P_{r_i}(x) + P_{f_i}(x)}dx\\
=&\int_{x} P_{r_i}(x) \log \frac{P_{r_i}(x)}{(P_{r_i}(x) + P_{f_i}(x))/2} \\
&+ P_{f_i}(x) \log \frac{P_{f_i}(x)}{(P_{r_i}(x) + P_{f_i}(x))/2}dx-(w_i + 1)\log 2\\
=&KL(P_{r_i}\parallel \frac{P_{r_i} + P_{f_i}}{2})+KL(P_{f_i}\parallel \frac{P_{r_i} + P_{f_i}}{2})\\
&-(w_i + 1)\log 2\\
=&-(w_i + 1)\log 2 + 2JSD(P_{r_i}\parallel P_{f_i})
\enspace.
\end{align*}
Since we already have the optimal solution for all clusters as given in Eq.~\ref{D-opt}, we can say that for each cluster, we minimize the JS divergence between real data and fake data. When summed up, the optimal remains since each term is independent with others.
\end{proof}

\subsection{Derivation for $\phi_i^*$}
\label{phi opt proof}
This result can be easily obtained by solving the optimization problem:
\begin{align*}
\begin{aligned}
& \underset{\phi}{\text{max}}
& & \mathop{\mathbb{E}} \limits_{x\sim P_{r_i}} \log \phi_i 
\enspace, \\
& \text{s.t.}
& & \sum_{i=1}^{K}\phi_i=1
\enspace.
\end{aligned}
\end{align*}
We can derive that
\begin{align*}
\phi_i = \frac{\sum_{x} P(z|x, \theta^{old})}{\sum_{z} \sum_{x} P(z|x, \theta^{old})} = \frac{N_i}{N} \enspace.
\end{align*}

\subsection{Results on all 40 CelebA attributes}
\label{emgan-celeba-results}
Here in Tab.~\ref{celeba-attr} we show the experiment results of GAN-EM on all 40 CelebA attributes with both unsupervised and semi-supervised settings. We bold top three results for unsupervised clustering and semi-supervised classification respectively.
\begin{table*}
\centering
\caption{Error rates of GAN-EM on all 40 CelebA attributes}
\hspace*{-3cm}
\begin{tabular}{l|ll|l|l|ll}
\hline
Attributes        & Unsupervised & \begin{tabular}[c]{@{}l@{}}Semi-supervised\\ (100 labels)\end{tabular} &  & Attributes            & Unsupervised & \begin{tabular}[c]{@{}l@{}}Semi-supervised\\ (100 labels)\end{tabular} \\ \hline
Goatee            & 0.4103       & 0.2218                                                                 &  & Bald                  & 0.4855       & 0.4001    \\
Narrow\_Eyes      & 0.3308       & 0.1757                                                                 &  & Arched\_Eyebrows      & 0.4447       & 0.2993          \\
Bangs             & 0.3592       & 0.1506                                                                 &  & Wavy\_Hair            & 0.3376       & 0.2657          \\
Gray\_Hair        & 0.4538       & 0.1849                                                                 &  & Mouth\_Slightly\_Open & 0.4272       & 0.1895          \\
Big\_Lips         & 0.4804       & 0.1767                                                                 &  & Young                 & 0.4557       & 0.3216          \\
Heavy\_Makeup     & 0.2366       & 0.2218                                                                 &  & No\_Beard             & 0.4869       & 0.4315          \\
Attractive        & 0.4749       & 0.3736                                                                 &  & Pointy\_Nose          & 0.3426       & 0.1847          \\
Bags\_Under\_Eyes & 0.4468       & 0.3545                                                                 &  & Bushy\_Eyebrows       & 0.4447       & 0.3592          \\
High\_Cheekbones  & 0.4455       & 0.2637                                                                 &  & Double\_Chin          & 0.4546       & 0.3294          \\
Oval\_Face        & 0.4858       & 0.2583                                                                 &  & gender                & 0.3683       & 0.1446          \\
Rosy\_Cheeks      & 0.3518       & 0.2891                                                                 &  & hat                   & 0.2941       & 0.1598          \\
Sideburns         & 0.4726       & 0.2507                                                                 &  & glass                 & 0.2983       & 0.1662          \\
Mustache          & 0.4539       & 0.2947                                                                 &  & Male                  & 0.4763       & 0.2746          \\
Brown\_Hair       & 0.4463       & 0.4362                                                                 &  & Receding\_Hairline    & 0.4847       & 0.4138          \\
Pale\_Skin        & 0.4736       & 0.4457                                                                 &  & Wearing\_Necklace     & 0.4758       & 0.2755          \\
Chubby            & 0.4478       & 0.3242                                                                 &  & Wearing\_Necktie      & 0.4805       & 0.4157          \\
Big\_Nose         & 0.4695       & 0.4337                                                                 &  & Wearing\_Lipstick     & 0.4141       & 0.1601          \\
Blurry            & 0.3045       & 0.2188                                                                 &  & Straight\_Hair        & 0.4687       & 0.3102          \\
Black\_Hair       & 0.4555       & 0.3949                                                                 &  & Wearing\_Earrings     & 0.4393       & 0.3686          \\
Blond\_Hair       & 0.3145       & 0.3057                                                                 &  & 5\_o\_Clock\_Shadow   & 0.4435       & 0.2822          \\ \hline
\end{tabular}
\label{celeba-attr}
\end{table*}

\section{Implementation details}
\label{imp-detail}
\subsection{MNIST}
MNIST dataset has 60,000 training samples, among which we assign 10,000 samples as validation data. The images are $28\times 28$ pixels of which the values are normalized to [0, 1]. The input of the generator is 72 dimension (62 dimension random noise and 10 dimension conditional class label)The generator has one fully connected layer with 6272 hidden units ($128\times 7\times 7$) followed by 2 transpose convolutional layers. The first transpose convolutional layer has 64 filters of which the size is $4\times 4$. The convolutional operation is stride 2 with 1 zero padding. The second convoluational layer uses the same structure with the first one except that the number of filters is 1. Then the output of the generator is in size $28\times 28$ which is the same as the real image size. The discriminator also has 3 convolutional layers with 64, 128, 256 filters respectively. All of these 3 layers uses kernel size 4, stride 2 and zero padding 1. Then 2 fully connected layers are used with 1024 and 11 hidden units respectively.  E-net shares the same structure with the discriminator except that the number of units in last layer is 10. Tab.~\ref{mnist-net} describes the network structure in detail.
\begin{table*}[h!]
\centering
\caption{GAN structure for MNIST.}
\label{mnist-net}
\begin{tabular}{l|l|l}
\hline
Generator & Discriminator & E-net\\ \hline
Input 72 dim vector & Input $28\times 28$ & Input $28\times 28$\\
FC ($128\times 7\times 7$)&$4\times 4$ Conv, f:64  s:2  d:1&$4\times 4$ Conv, f:64  s:2  d:1\\
BN lRelu(0.2)&BN lRelu(0.2)&BN lRelu(0.2)\\
$4\times 4$ Deconv, f:64  s:2  d:1&$4\times 4$ Conv, f:128  s:2  d:1&$4\times 4$ Conv, f:128  s:2  d:1\\
BN lRelu(0.2)&BN lRelu(0.2)&BN lRelu(0.2)\\
$4\times 4$ Deconv, f:1  s:2  d:1&$4\times 4$ Conv, f:256  s:2  d:1&$4\times 4$ Conv, f:256  s:2  d:1\\
&BN lRelu(0.2)&BN lRelu(0.2)\\
&FC(1024), BN lRelu(0.2)&FC(1024), BN lRelu(0.2)\\
&FC(11)&FC(10)\\ \hline
\end{tabular}
\begin{tabular}{l}
BN: batch normalization, FC: fully connected layer. f: filter number. s: stride size. d: padding size
\end{tabular}
\end{table*}

We apply RMSprop optimizer to all these 3 networks with learning rate 0.0002 (decay rate: 0.98). The random noise of generator is in uniform distribution. In each M-step, there are 5 epoches with a minibatch size of 70 for both the generated batch and the real samples batch. We use a same update frequency for generator and discriminator. For E-step, we generate samples using well trained generator with batch size of 200, then we apply 1000 iterations to update E-net.

For semi-supervised classification, we add the supervision to both E-net. At the end of E-step, we train the E-net by labeled data until the prediction accuracy given by E-net is 100\%. Then we feed the real data to E-net to obtain $P(c|x_{real})$ which is the same as unsupervised clustering.

\subsection{SVHN}
Similar to MNIST, SVHN is also a digits recognition datasets with about 53,000 training images ($32\time 32$ pixels). Since the images of SVHN are all from the real world and many of them are blurry, the recognition problem is much more difficult than MNIST which is preprocessed to grey-scale images. We normalize the images to [-1, 1]. We use almost the same network structure with MNIST with only a slight difference. The details of GAN are in Tab.~\ref{svhn-net}.
\begin{table*}[h!]
\centering
\caption{GAN structure for SVHN.}
\label{svhn-net}
\begin{tabular}{l|l|l}
\hline
Generator & Discriminator & E-net\\ \hline
Input 72 dim vector & Input $28\times 28$ & Input $28\times 28$\\
FC ($128\times 4\times 4$)&$4\times 4$ Conv, f:64  s:2  d:1&$4\times 4$ Conv, f:64  s:2  d:1\\
BN lRelu(0.2)&BN lRelu(0.2)&BN lRelu(0.2)\\
$4\times 4$ Deconv, f:64  s:2  d:1&$4\times 4$ Conv, f:128  s:2  d:1&$4\times 4$ Conv, f:128  s:2  d:1\\
BN lRelu(0.2)&BN lRelu(0.2)&BN lRelu(0.2)\\
$4\times 4$ Deconv, f:32  s:2  d:1&$4\times 4$ Conv, f:256  s:2  d:1&$4\times 4$ Conv, f:256  s:2  d:1\\
BN lRelu(0.2)&BN lRelu(0.2)&BN lRelu(0.2)\\
$4\times 4$ Deconv, f:3  s:2  d:1&FC(1024), BN lRelu(0.2)&FC(1024), BN lRelu(0.2)\\
&FC(11)&FC(10)\\ \hline
\end{tabular}
\begin{tabular}{l}
BN: batch normalization, FC: fully connected layer. f: filter number. s: stride size. d: padding size
\end{tabular}
\end{table*}

For semi supervision, we use the same strategies with MNIST which is discussed above.

\subsection{CelebA}
\label{celeba-appendix}
CelebA dataset is a large-scale human face dataset with more than 200k images. Each image is annotated by 40 binary attributes. For example, for gender attribute, we first select all 12k male images, then we select the same amount of female images. The selection of female images is based on such principle: all other attributes except for gender should maintain as much purity as possible. Since we are unable to guarantee all the other attributes are 100\% pure, we regard those impure attributes as noise. The selected images are cropped to $64\times 64$ pixels and are normalized to [-1, 1]. We apply the same strategy for the other attributes. The network details are in Tab.~\ref{celeba-net}.
\begin{table*}[h!]
\centering
\caption{GAN structure for CelebA.}
\label{celeba-net}
\begin{tabular}{l|l|l}
\hline
Generator & Discriminator & E-net\\ \hline
Input 300 dim vector & Input $64\times 64$ & Input $64\times 64$\\
$4\times 4$ Deconv, f:1024 s:1 d:0 &$4\times 4$ Conv, f:128  s:2  d:1&$4\times 4$ Conv, f:128  s:2  d:1\\
BN lRelu(0.2)&BN lRelu(0.2)&BN lRelu(0.2)\\
$4\times 4$ Deconv, f:512  s:1  d:0&$4\times 4$ Conv, f:256  s:2  d:1&$4\times 4$ Conv, f:256  s:2  d:1\\
BN lRelu(0.2)&BN lRelu(0.2)&BN lRelu(0.2)\\
$4\times 4$ Deconv, f:256  s:1  d:0&$4\times 4$ Conv, f:512  s:2  d:1&$4\times 4$ Conv, f:512  s:2  d:1\\
BN lRelu(0.2)&BN lRelu(0.2)&BN lRelu(0.2)\\
$4\times 4$ Deconv, f:128  s:1  d:0&$4\times 4$ Conv, f:1024  s:2  d:1&$4\times 4$ Conv, f:1024  s:2  d:1\\
BN lRelu(0.2)&BN lRelu(0.2)&BN lRelu(0.2)\\
$4\times 4$ Deconv, f:3  s:1  d:0&$4\times 4$ Conv, f:3  s:1  d:0&$4\times 4$ Conv, f:3  s:1  d:0\\ 
&BN lRelu(0.2)&BN lRelu(0.2)\\ 
&FC(3)&FC(2)\\ \hline
\end{tabular}
\begin{tabular}{l}
BN: batch normalization, FC: fully connected layer. f: filter number. s: stride size. d: padding size
\end{tabular}
\end{table*}

\end{appendices}

\end{document}